\documentclass[preprint,12pt]{elsarticle}

\usepackage[latin1]{inputenc}
\usepackage{amssymb}
\usepackage{amsmath}
\usepackage{graphicx}
\usepackage{bm}

\usepackage{color}
\usepackage{array}

\usepackage[numbers]{natbib}

\journal{Neurocomputing}

\begin{document}

\begin{frontmatter}

\title{Dimensionality reduction for time series data}

\author[oxford1]{Diego Vidaurre}
\ead{diego.vidaurre@ohba.ox.ac.uk}
\author[oxford2]{Iead Rezek} 
\ead{iead@rezek.info}
\author[oxford3]{Samuel J Harrison}
\ead{samuel.harrison@balliol.ox.ac.uk}
\author[oxford3]{Stephen M Smith} 
\ead{steve@fmrib.ox.ac.uk}
\author[oxford1]{Mark W Woolrich} 
\ead{mark.woolrich@ohba.ox.ac.uk}

\address[oxford1]{Oxford Centre for Human Brain Activity (OHBA)}
\address[oxford2]{Department of Engineering Science}
\address[oxford3]{Oxford Centre for Functional MRI of the Brain (FMRIB), \\ University of Oxford, Oxford, UK}

\begin{abstract} 
Despite the fact that they do not consider the temporal nature of data,  
classic dimensionality reduction techniques, such as PCA, are widely applied to time series data. 
In this paper, we introduce a factor decomposition specific for time series that builds upon the Bayesian multivariate autoregressive model 
and hence evades the assumption that data points are mutually independent.  
The key is to find a low-rank estimation of the autoregressive matrices.
As in the probabilistic version of other factor models, this induces a latent low-dimensional representation of the original data.    
We discuss some possible generalisations and alternatives, 
with the most relevant being a technique for simultaneous smoothing and dimensionality reduction.    
To illustrate the potential applications, we apply the model on a synthetic data set and  different types of neuroimaging data (EEG and ECoG).
\end{abstract}

\end{frontmatter}

\section{Introduction}

In this paper, we introduce a novel low-rank factorisation based in the multivariate autoregressive (MAR) model \cite{box11,penny02}.
The MAR model characterises the behaviour of time series by linear historical interactions between the $N$ variables or channels. 
The central parameter of a MAR model is the order or length of the linear interaction, denoted as $P$. 
A MAR model is thus composed by $P$ matrices of autoregressive coefficients of size $N \times N$. 
We propose to carry out a low-rank approximation of these matrices so that the 
effective number of parameters drops from $P N^2$ to $P N Q$, being $Q<N$.
Such factorisation permits to express the original $N$-dimensional signal in a lower $Q$-dimensional space. 
We name the model low-rank MAR (LR-MAR).

Therefore, unlike other probabilistic latent variable models,
such as principal component analysis (PCA) \citep{bishop99},
independent component analysis (ICA) \cite{beckmann04} and
canonical correlation analysis (CCA) \citep{wang07,fujiwara13},
LR-MAR considers the temporal nature of the data.
Avoiding the independence assumption of the data, 
we expect LR-MAR to outperform the other probabilistic latent variable models when dealing with time series data. 
LR-MAR is a less complex model than MAR (in the sense of the number of parameters) and, then, 
it can also be considered as an alternative to MAR to prevent overfitting in high-dimensional scenarios. 

Additionally, we generalise the proposed model by modifying the autoregression problem so that
multiple lags are simultaneously estimated as a linear function of the previous data points. 
This approach compactly accomplishes dimensionality reduction and data smoothing in one go and has a connection with CCA.
CCA finds linear combinations of two (or more) groups of random variables with the maximum correlation with each other.
These combinations can be used to analyse the common variability of the input variables.   
Because of this connection, we denote this approach as windowed CCA (wCCA).



There exist some models in the literature that are related to LR-MAR.
The recursive PCA algorithm, for example, is an alternative for dimensionality reduction of time series data,
based on carrying out PCA over a moving window \cite{li00}. 
The philosophy of our approach is different in the sense that LR-MAR is a decomposition that considers time dynamics
instead of a decomposition that changes over time,
thus allowing for a compact representation of the entire time series. 
 
In a Kalman smoother (KS) \cite{bishop06}, 
the observations are conditionally independent given the state of a latent signal.
Choosing its dimension to be lower than $N$, 
the latent signal can be regarded as a low-dimensionality representation of the observed signal.
However, marginalising out the latent variable, the predictive distribution of the observed signal given previous observations
has a more complex form than in the MAR model \cite{bealthesis}.
Also, it can be shown that any MAR model can be expressed as an equivalent KS model. 
In this case, however, the required parametrisation is such that the 
 latent signal of the resulting KS model has a higher dimension than the observed signal. 
Among other inference methods, variational inference procedures are available for KS models \cite{barber06}.

KS models have been extended beyond linearity in variety of ways. 
For example, \cite{fox08} consider a switching KS, 
where the system dynamics and output function regime depend on time and can be chosen from a finite number of models.  
This number is however not specified a priori and is elicited from the data. 
A different approach is to consider nonlinear dynamics and/or nonlinear output functions;
see e.g. \cite{kitagawa96,wang05,schon05}.
In this paper, however, we stay within linearity, which usually provides more interpretable and practical models.

We give a Bayesian formulation of both LR-MAR and wCCA, using conjugate Gaussian priors on the autoregression coefficients,
so that, in addition to further controlling the effective complexity of the model, 
we achieve sparsity by means of the automatic relevance determination principle \cite{bishop06}.  
The LR-MAR model can also be considered as a regularised version of the standard MAR model. 
This also serves an interpretability purpose, as it helps to identify those variables that are mostly noise in the original data. 
Conjugate priors allow for computationally attractive VB inference.
The evaluation of the variational free is a natural tool for model selection, i.e. 
for finding the appropriate autoregressive order and number of latent components.


The rest of the paper is organised as follows. 
Section 2 introduces the model.
Section 3 provides the VB equations and the expression of the free energy. 
Section 4 discusses some extensions and generalisations. 
Section 5 illustrates the model performance over some data sets. 
Finally, in Section 6, we draw some conclusions.

\section{Linear MAR-based decomposition}
\label{basic}

Let $\boldsymbol{y}_t \in \mathbb{R}^N$ be a column vector representing the multi-channel source signal at time $t$ and $\boldsymbol{y}'_t$ its transpose.
We denote the entire source signal as $\boldsymbol{Y} \in \mathbb{R}^{T \times N}$ and
assume an autoregression model of order $P$ with autoregression parameters $\boldsymbol{B}_i$, $i=1,...,P$.
We assume centered data.
In this paper, we propose a $Q$-rank approximation of the usual MAR model, given by



\vspace{-0.3cm}
\begin{equation*}
\boldsymbol{y}'_t  =  \sum_{i=1}^P \boldsymbol{y}'_{t-i}  \boldsymbol{B}_i +  \boldsymbol\epsilon = 
\sum_{i=1}^P \boldsymbol{y}'_{t-i}  \boldsymbol{W}_i \boldsymbol{V} +  \boldsymbol\epsilon,
\end{equation*}
\vspace{-0.3cm}

\noindent  
where $\boldsymbol{W}_i$ and $\boldsymbol{V}$ are, respectively, $N \times Q$ and $Q \times N$ dimensional matrices
and $\boldsymbol\epsilon$ is white Gaussian noise.
We define additional hidden variables $\boldsymbol{z}_t   \in \mathbb{R}^Q$ that encode a low-rank representation of the signal.
Then,

\vspace{-0.5cm}
\begin{equation}
\label{zy}
\boldsymbol{z}'_t \, | \, \boldsymbol{y}_{t-P},...,\boldsymbol{y}_{t-1}, \boldsymbol{W}_1, ..., \boldsymbol{W}_P  \quad \sim \quad \mathcal{N}\Big( \sum_{i=1}^P \boldsymbol{y}'_{t-i}  \boldsymbol{W}_i, \boldsymbol{I}_Q \Big),
\end{equation}
\begin{equation*}
\boldsymbol{y}'_t \, | \, \boldsymbol{z}_t ,\boldsymbol{V} , \boldsymbol\Omega  \quad \sim \quad \mathcal{N}\Big( \boldsymbol{z}'_{t}  \boldsymbol{V} , \boldsymbol\Omega \Big),
\end{equation*}
\vspace{-0.3cm}

\noindent
where $\boldsymbol\Omega$ is a diagonal covariance matrix, with diagonal elements
$\Omega^{-1}_{nn} \, \sim \, \mathcal{G}(\iota ,a_n)$.   
We set the covariance of $\boldsymbol{z}'_t$ to be the identity matrix $\boldsymbol{I}_Q$ for identifiability purposes. 
Using the semicolon to indicate vertical concatenation, 
we shall denote $\boldsymbol{Z} = (\boldsymbol{z}'_1; ... ; \boldsymbol{z}'_T) \in \mathbb{R}^{T \times Q}$.

To be robust to the case when some channels are just noise, 
and to be able to select the lags of interest,
we set automatic relevance determination (ARD) priors on the rows of $\boldsymbol{W}_i$, controlled by parameters 
$\alpha^2_{i1},...,.\alpha^{2}_{iN}$.
In order to control the dimensionality of the latent space, we also impose ARD priors on the rows of $\boldsymbol{V}$, 
controlled by a parameter $\boldsymbol\gamma^2 = (\gamma^2_1,...,.\gamma^2_Q)$.
Thus, for $j=1,...,Q$, $n=1,...,N$ and $i=1,...,P$, we have

\vspace{-0.3cm}
\begin{equation*}
W_{i_{nj}} \sim \mathcal{N} \big( 0, \alpha_{in}^{-2}  \big), \quad \quad \quad \quad 
V_{jn} \sim \mathcal{N} \big(  0, \gamma_{j}^{-2}  \big).
\end{equation*}	
\vspace{-0.3cm}

Finally, we set Gamma distributions on the ARD precisions, 
$\alpha_{in}^2 \, \sim \, \mathcal{G}(\kappa ,b_n )$ 
and $\gamma_{j}^2 \, \sim \, \mathcal{G}(\nu ,c_j)$.

\begin{center}
\begin{figure}[t!]
\hspace{2.4cm} 
\includegraphics[width=3.25in, totalheight=2.5in]{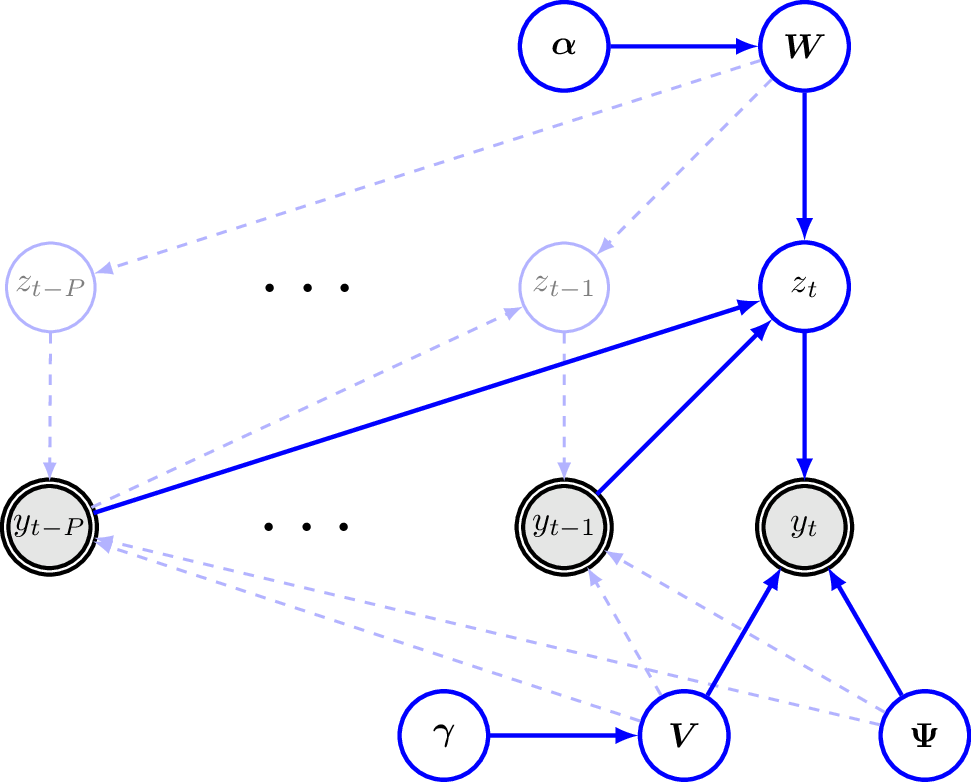}
\caption{Graphical representation of the Bayesian hierarchy.} 
\label{graphrepr}
\end{figure}
\end{center}

Figure \ref{graphrepr} shows the model graphically. 

\vspace{-0.9cm}

\section{Variational parameter inference}

In this section, we use VB to estimate the parameters of the model.
For the observation model, we approximate the posterior distribution of the parameters, 
$Pr(\boldsymbol{Z}, \boldsymbol{W},\boldsymbol{V}, \boldsymbol{\Omega}^{-1}, \boldsymbol\alpha^{-2},\ \boldsymbol{\gamma}^{-2} \mid \boldsymbol{Y})$,  
by a variational distribution $F(\boldsymbol{Z}, \boldsymbol{W},\boldsymbol{V}, \boldsymbol{\Omega}^{-1},\boldsymbol{\alpha}^{-2},\boldsymbol{\gamma}^{-2})$, 
which factorizes as follows 

\begin{eqnarray*}
Pr(\boldsymbol{Z}, \boldsymbol{W},\boldsymbol{V}, \boldsymbol{\Omega}^{-1}, \boldsymbol\alpha^{-2},\ \boldsymbol{\gamma}^{-2} \mid \boldsymbol{Y}) \approx
\quad\quad\quad\quad\quad\quad\quad\quad\quad\quad\quad  \quad\quad\quad \\
F(\boldsymbol{Z}, \boldsymbol{W},\boldsymbol{V}, \boldsymbol{\Omega}^{-1},\boldsymbol{\alpha}^{-2},\boldsymbol{\gamma}^{-2}) = 
F(\boldsymbol{Z}) F(\boldsymbol{W},\boldsymbol{\alpha}^{-2})
F(\boldsymbol{V},\boldsymbol{\Omega}^{-1},\boldsymbol{\gamma}^{-2}).
\end{eqnarray*}

For ease of inference, we also make the assumptions  
$F(\boldsymbol{W},\boldsymbol\alpha^{-2}) = F(\boldsymbol{W}) F(\boldsymbol{\alpha}^{-2})$ and 
$F(\boldsymbol{V},\boldsymbol{\Omega}^{-1},\boldsymbol{\gamma}^{-2}) = F(\boldsymbol{V}) F(\boldsymbol{\Omega}^{-1},\boldsymbol{\gamma}^{-1})$.
A factorisation between $\boldsymbol{\Omega}^{-1}$ and $\boldsymbol{\gamma}^{-1}$  comes naturally due to the DAG structure. %
$F(\boldsymbol{Z})$ also factorizes with no need for the assumption of $\prod_{t=P+1}^T F(\boldsymbol{z}_t)$.

Considering $F(\boldsymbol{Z})$ as a hidden unknown, 
the variational approximation for the observational model comprises itself a variational pair of E-step and M-step. 
The E-step computes

\begin{equation}
\label{z}
F(\boldsymbol{z}_t) = \mathcal{N}(\boldsymbol{z}_t; \bar{\boldsymbol{z}}_t,\boldsymbol{S}_{\boldsymbol{z}_t})
\end{equation}

\noindent
with

\begin{equation*}
\boldsymbol{S}_{\boldsymbol{z}_t} =  \Big( \boldsymbol{I}_Q + E[\boldsymbol{V} \boldsymbol{\Omega}^{-1} \boldsymbol{V}']\Big)^{-1}, \quad
\bar{\boldsymbol{z}_t} =  \boldsymbol{S}_{\boldsymbol{z}_t} 
\bigg(  \sum_{i=1}^P  \bar{\boldsymbol{W}}'_i \boldsymbol{y}_{t-i} + 
\bar{\boldsymbol{V}} \bar{\boldsymbol{\Omega}}^{-1} \boldsymbol{y}_t  \bigg),
\end{equation*}

\noindent
where $E[\boldsymbol{V} \boldsymbol{\Omega}^{-1} \boldsymbol{V}'] = \bar{\boldsymbol{V}} \bar{\boldsymbol\Omega}^{-1} \bar{\boldsymbol{V}}' + \sum_{n=1}^N \bar{\Omega}_{nn}^{-1} \boldsymbol{S}_{\boldsymbol{V}_{n}}$, 
expectations with respect to $F(\cdot)$  are denoted with an upper bar (e.g., $\bar{\boldsymbol{W}}_i$ and $\bar{\boldsymbol{V}}$) and 
$\boldsymbol{S}_{\boldsymbol{V}_{n}}$ denotes the $N \times N$ covariance matrix for  the $n$-th 
column of the matrix $\boldsymbol{V}$. 

We shall denote 
$\boldsymbol{W} = [\boldsymbol{W}_1; ...; \boldsymbol{W}_P] \in \mathbb{R}^{NP \times Q}$, containing the the autoregression coefficients for all lags.
We refer to the $j$-th column of $\boldsymbol{W}$ as $\boldsymbol{w}_j$.
Let $\boldsymbol{Y}^+ = [\boldsymbol{y}'_{P+1};...; \boldsymbol{y}'_{T}] \in \mathbb{R}^{ T-P\times N}$,
$\boldsymbol{Y}_i^- = [\boldsymbol{y}'_{P-i+1};...; \boldsymbol{y}'_{T-i}] \in \mathbb{R}^{ T-P\times N}$
and $\boldsymbol{Y}^- = [\boldsymbol{Y}_1^- ... \boldsymbol{Y}_P^- ] \in \mathbb{R}^{T-P \times NP}$,
so that, assuming $\boldsymbol\Omega^{-1}$ to be known, the log likelihood of the observed time series is given by 

\begin{equation*}
- \frac{(T-P)N}{2} \log (2\pi) - \frac{(T-P)}{2} \log | \boldsymbol\Omega^{-1} | -
\mathrm{tr} \Big( (\boldsymbol{Y}^+ - \boldsymbol{Y}^- \boldsymbol{W} \boldsymbol{V}) \boldsymbol\Omega^{-1} (\boldsymbol{Y}^+ - \boldsymbol{Y}^- \boldsymbol{W} \boldsymbol{V})' \Big)
\end{equation*}

\noindent
Similarly, we denote $\boldsymbol\alpha^{-2} = (\alpha_{11}^{-2},...,\alpha_{1N}^{-2},...,\alpha_{P1}^{-2},...,\alpha_{PN}^{^-2})$ to contain the concatenated prior parameters for $\boldsymbol{W}$.
Also, $\boldsymbol{Z}_{\cdot j}$ is the $j$-th column of $\boldsymbol{Z}$.

Thanks to the identity covariance matrix assumption in Equation (\ref{zy}), we can without further assumptions factorize 
$F(\boldsymbol{W}) = \prod_{j=1}^Q F(\boldsymbol{w}_j)$.
For each factor, we have

\begin{equation}
\label{w-sparse-fact}
F(\boldsymbol{w}_j)= \mathcal{N}(\boldsymbol{w}_j; \bar{\boldsymbol{w}}_j,\boldsymbol{S}_{\boldsymbol{w}_j})
\end{equation}

\noindent
with 

\begin{equation*}
\boldsymbol{S}_{\boldsymbol{w}_j} = \Big( \mathrm{diag}(\bar{\boldsymbol\alpha}^{-2}) + \boldsymbol{Y}^{-'}  \boldsymbol{Y}^- \Big)^{-1} , 
\quad \quad
 \bar{\boldsymbol{w}}_j  =  \boldsymbol{S}_{\boldsymbol{w}_j}
\boldsymbol{Y}^{-'}    \bar{\boldsymbol{Z}}_{\cdot j}.
\end{equation*}

For $\alpha^{-2}_{in}$, we have a Gamma distribution 

\begin{equation}
\label{alpha}
F(\alpha^{-2}_{in}) = \mathcal{G} \Big(\alpha^{-2}_{in}; \,\, \tilde{\kappa} , \tilde{b}_{in} \Big)
\end{equation}

\noindent
with shape $\tilde{\kappa} =  \kappa + \frac{Q}{2}$ and rate $\tilde{b}_{in} =  \frac{1}{2} E[\boldsymbol{W}'_{i_{n \cdot}} \boldsymbol{W}_{i_{n \cdot}}] + b_{in}$.

For $\boldsymbol{V}$, we can also factorize $F(\boldsymbol{V}) = \prod_{n=1}^N F(\boldsymbol{v}^{n})$, so that we have

\begin{equation}
\label{v-sparse-fact}
F(\boldsymbol{v}_n)= \mathcal{N}(\boldsymbol{v}_n; \,\, \bar{\boldsymbol{v}}_n,\boldsymbol{S}_{\boldsymbol{v}_n})
\end{equation}

\noindent
with 

\begin{equation*}
\boldsymbol{S}_{\boldsymbol{v}_n} = \Big( \mathrm{diag}( \bar{\boldsymbol\gamma}^{-2}  ) + \bar{\Omega}_{nn}^{-1}  E[\boldsymbol{Z}'  \boldsymbol{Z}]  \Big)^{-1} , \quad\quad
  \bar{\boldsymbol{v}}_n  = \bar{\Omega}_{nn}^{-1} \boldsymbol{S}_{\boldsymbol{v}_n}
\bar{\boldsymbol{Z}}'  \boldsymbol{Y}^+_{\cdot n}.
\end{equation*}


For each element $\Omega_{nn}^{-1}$, we have a Gamma distribution given by

\begin{equation}
\label{omega-diag}
F(\Omega_{nn}^{-1}) = \mathcal{G}\Big(\Omega_{nn}^{-1}; \,\,\tilde{\iota}, \tilde{a}_n \Big)
\end{equation}

\noindent
with shape $\tilde{\iota} = \iota + \frac{T-P}{2}$ and rate $\tilde{a}_n = \frac{1}{2} \Big( \boldsymbol{Y}^{+'}_{\cdot n} \boldsymbol{Y}^{+}_{\cdot n} +  
E[\boldsymbol{v}_n'  \boldsymbol{Z}' \boldsymbol{Z}  \boldsymbol{v}_n]    - 2  \boldsymbol{Y}^{+'}_{\cdot n}  \bar{\boldsymbol{Z}} \bar{\boldsymbol{v}}_n \Big)+ a_n$.

For $\gamma^{-2}_j$, we have a Gamma distribution 

\begin{equation}
\label{gamma}
F(\gamma^{-2}_j) = \mathcal{G} \Big(\gamma^{-2}_j; \,\, \tilde{\nu}, \tilde{c}_j \Big)
\end{equation}

\noindent
with shape $\tilde{\nu} = \nu + \frac{N}{2}$ and rate $\tilde{c}_j =  \frac{1}{2} E[\boldsymbol{V}'_{j \cdot} \boldsymbol{V}_{j \cdot}] + c_j$.





In summary, the algorithm alternates the computation of functionals (\ref{z}), (\ref{w-sparse-fact}), (\ref{alpha}),
(\ref{v-sparse-fact}), (\ref{omega-diag}) and (\ref{gamma})
The ordering in the computation of the functionals can be driven by the DAG structure but it is not crucial. 

\noindent
The marginal predictive distribution is then given by

\begin{equation}
\boldsymbol{y}_t  \,\, \sim \,\, \mathcal{N}\bigg(
\bar{\boldsymbol{V}}' \sum_{k=1}^K  \sum_{i=1}^P \bar{x}_{tk} \bar{\boldsymbol{W}}^{(k)'}_i \boldsymbol{y}_{t-i} \,\,,\,\,
\bar{\boldsymbol\Omega} + E[ \boldsymbol{V}'   \boldsymbol{V}]  \bigg),
\end{equation}

The derivation of the free energy, useful for monitoring model selection purposes, is given in the Appendix.

\vspace{1cm}
\section{Extending to multiple output lags}
\label{generalizations}

In this section, we present a generalisation of the above method where the autoregression problem is set so that the response variable contains several lags. 
Building upon this idea, we introduce wCCA, a useful variation for simultaneous data smoothing and dimensionality reduction based on CCA.
This is a compact alternative to common practice approaches, that perform dimensionality reduction and smoothing in two separate steps,
and is particularly useful for high-dimensional setting, where interpretability is a main concern.
The main advantage is that we can apply standard Bayesian methodology for model selection in a principled way, 
whereas parameter tuning when one performs dimensionality reduction and smoothing in separated steps is typically guided by heuristics and rules-of-thumb.  

The generalisation is done by extending $\boldsymbol{Y}^+$ so that each row contains more than one lag. This way, we redefine
$\boldsymbol{Y}^+ = [\boldsymbol{Y}_1^+ ... \boldsymbol{Y}_L^+ ] \in \mathbb{R}^{(T-P-L+1) \times NL}$, where
the $\boldsymbol{Y}_l^+ \in \mathbb{R}^{(T-P-L+1) \times N}$ matrices have rows $\boldsymbol{y}'_{P+l},..., \boldsymbol{y}'_{T-L+l}$.
$\boldsymbol{Y}^-$ is defined as before, but removing the last $L$ rows.
Hence, $\boldsymbol{Y}^-$ now has dimension  $\mathbb{R}^{(T-P-L+1) \times NP}$.
Then, $\boldsymbol{z}_t$ is the latent variable that
corresponds to the low-rank estimation of $\boldsymbol{y}_t,...,\boldsymbol{y}_{t+L-1}$ 
using $\boldsymbol{y}_{t-1},...,\boldsymbol{y}_{t-P}$ as inputs. 

One possibility is to extend the LR-MAR model to

\begin{equation}
\label{yy}
\boldsymbol{Y}^+ = \boldsymbol{Y}^- \boldsymbol{W} \boldsymbol{V} + \boldsymbol\epsilon,
\end{equation}

\noindent
where $\boldsymbol{V}$ has now dimension $Q \times NL$ and 
$\boldsymbol\epsilon \in \mathbb{R}^{T-L-P \times NL}$ is Gaussian noise.  
This formulation can be regarded as a (Bayesian) partial least squares model \cite{vidaurre13} 
built from the autoregression setup. 

The updating equations can be straightforwardly adapted to this case.
For example, the estimation of the sufficient parameters of the latent variable is

\begin{equation*}
\boldsymbol{S}_{\boldsymbol{z}_t} =  \Big( \boldsymbol{I}_Q + E[\boldsymbol{V} \boldsymbol{\Omega}^{-1} \boldsymbol{V}']\Big)^{-1}, \quad\quad\quad
\bar{\boldsymbol{z}_t} =  \boldsymbol{S}_{\boldsymbol{z}_t} 
\bigg(  \sum_{i=1}^P   \bar{\boldsymbol{W}}_i \boldsymbol{y}_{t-i} + 
\bar{\boldsymbol{V}} \bar{\boldsymbol{\Omega}}^{-1} \boldsymbol{y}^+_t  \bigg),
\end{equation*}

\noindent 
where $\boldsymbol{\Omega}$ is now $NL \times NL$-dimensional  
and $\boldsymbol{y}^+_t = [ \boldsymbol{y}_t \, ... \,\boldsymbol{y}_{t+L-1}] \in \mathbb{R}^{NL \times 1}$
is the $t$-th row of $\boldsymbol{Y}^+$.
This


More generally, if we abstract the time series data into two group of variables, an input $\boldsymbol{Y}^-$ 
and an output $\boldsymbol{Y}^+$, we can apply any low-rank regression methodology to obtain alternative dimensionality reduction techniques
specific for time series.

Considering $L=P$ for simplicity, we think that it is of particular interest to use (Bayesian) CCA \cite{wang07},
which treats $\boldsymbol{Y}^-$ and $\boldsymbol{Y}^+$ symmetrically.
This yields our proposed wCCA model, which can be modelled as

\begin{equation*}
\boldsymbol{Y}^- = \boldsymbol{Z} \boldsymbol{F} + \boldsymbol{E}_1 , \quad \quad \quad \quad
\boldsymbol{Y}^+ = \boldsymbol{Z} \boldsymbol{G} + \boldsymbol{E}_2,
\end{equation*}

\noindent
where $\boldsymbol{F}$ and $\boldsymbol{G}$ are $Q \times NP$ matrices and noise is assumed to be Gaussian.
The latent variable $\boldsymbol{Z}$ is Gaussian distributed, and ARD priors are applied over $\boldsymbol{F}$ and $\boldsymbol{G}$.
Therefore, we have that $\boldsymbol{z}_t$ represents the low-dimensional canonical correlation 
between data at time points $(t-P,...,T-1)$ and data at time points  $(T,...,L-1)$. 
Considering these two intervals as an effective smoothing window, 
this is effectively a Bayesian low-dimensional data smoothing approach. 

The parameters governing this model can be inferred using any procedure for solving the CCA problem. 
In this paper, we use the (Bayesian) formulation and inference procedure devised by \cite{fujiwara13}.

\section{Experiments}

In this section, we compare the proportion of variance explained by LR-MAR to PCA and ICA on synthetic time series data,
showing that considering the time structure of the data is useful to produce more informative and robust latent components. 
We also demonstrate the capacity of the method for extracting latent components 
that are meaningful in subsequent classification and regression tasks with real data, compared to PCA and ICA.
Furthermore, we show the performance of wCCA for simultaneous smoothing and dimensionality reduction.

\subsection{Synthetic simulations}

In order to demonstrate the performance of LR-MAR in a controlled scenario, 
we first generate 100 repetition of time courses with $T=4000$ and $N=12$ signals. 
Each signal is a weighted sum of some sinusoids plus Gaussian white noise. 
We consider 6 different sinusoids with different frequencies. 
For each signal, each sinusoid has probability $0.4$ to be included in the signal and, in case it is included, 
its weight is sampled from a Gamma distribution with both parameters equal to one.
The phase is set at random, independently for each signal. 

\begin{center}
\begin{figure}[h!]
\hspace{-1.4cm} 
\includegraphics[width=6.5in, totalheight=6.5in]{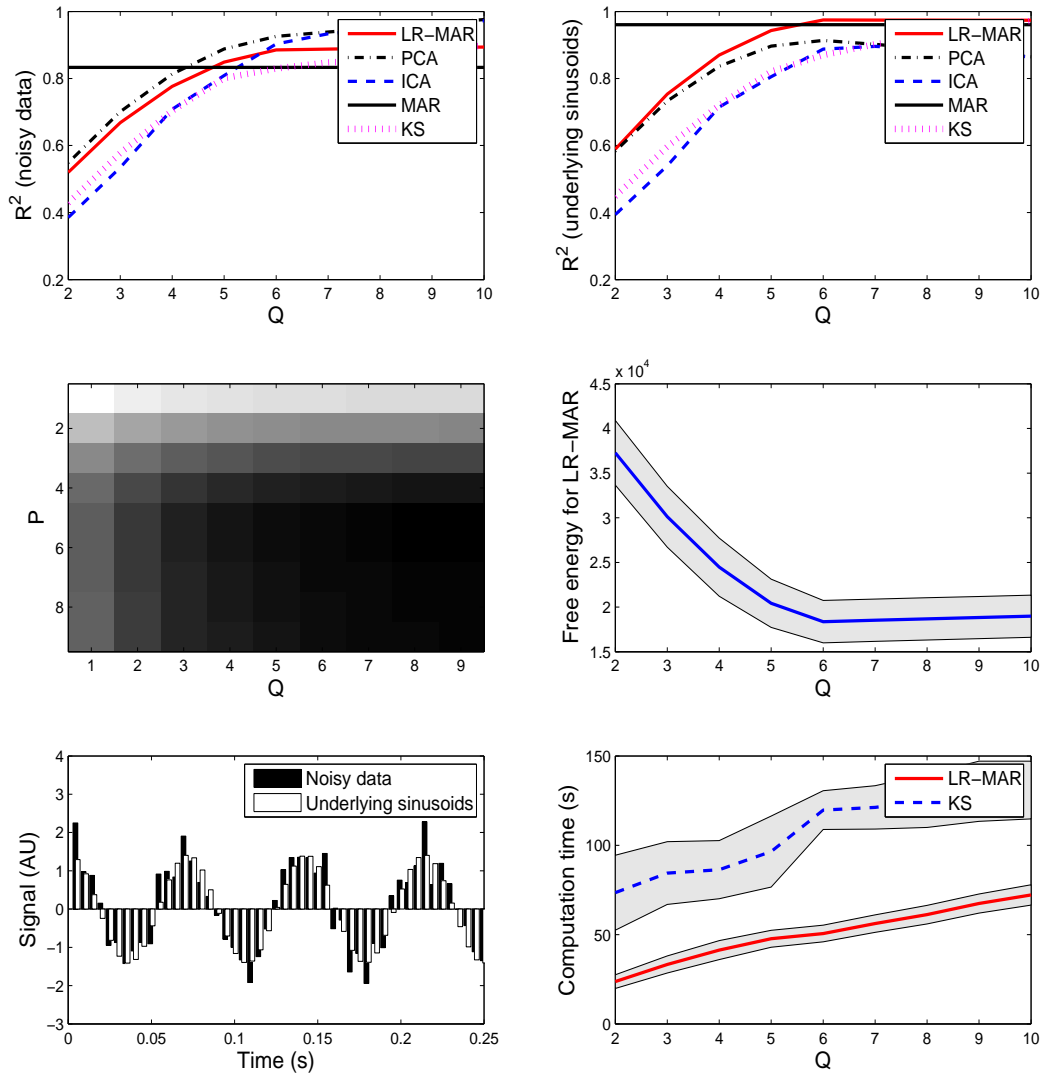}
\vspace{-1.5cm} 
\caption{Percentage of explained variance by the different methods for the noisy data (top left) and the underlying sinusoids (top right);
free energy $\mathcal{F}$ for LR-MAR as a function of $Q$ and $P$ (darker colours correspond to lower values, middle left);
evolution of the mean $\mathcal{F}$ ($\pm$ standard deviation) as a function of $Q$ for fixed $P=6$ (middle right);
an examplary signal before and after adding noise in arbitrary units (AU, left bottom);
the computation time of LR-MAR and KS as a function of $Q$ (right bottom). 
} 
\label{recon}
\end{figure}
\end{center}

The top panels of Figure \ref{recon} show the proportion of variance of the complete (noisy) data and the underlying sinusoids
explained by LR-MAR, PCA, ICA, MAR and KS for different values of $Q$. The MAR models were also computed with $P=6$.
Note that PCA and ICA, in order to compute the latent components at time $t$ (and reconstruct the signal from such latent components), 
use the signal at time $t$, whereas LR-MAR/MAR use the $P=6$ previous time points but not $t$. 
For the complete data, PCA explains most of the variance for $Q=10$ (explaining all of it for $Q=N$ by definition).
LR-MAR's performance to recover the signal is not much lower than PCA for the complete data and even better for the underlying sinusoids when $Q \to N$. 
ICA recovers less variance of the underlying sinusoids than PCA and LR-MAR, probably because ICA aims to decouple higher order moments.
KS is not far from LR-MAR in recovering the noisy data but does a worse job with the underlyings sinusoids.
For $Q \geq 6$, there is not a big difference between MAR and LR-MAR, suggesting that the average rank of the data (apart from noise) is 6,
which is  the number of different sinusoids used to generate the data sets. 
The middle left panel shows the evolution of the free energy $\mathcal{F}$ for LR-MAR as a function of $Q$, 
telling that $P=6,Q=6$ is the best choice, with little change for $Q>6$.
The middle right panel shows  $\mathcal{F}$ with more detail for $P=6$.
The bottom left panel shows an example of one generated signal superposed to the underlying sum of sinusoids.
The bottom right panel illustrates the computational cost in seconds for both LR-MAR and KS (both estimated using variational inference) as $Q$ grows, 
demonstrating a good, scalable computational efficiency for LR-MAR. 

These experiments show the good performance of LR-MAR when dealing with temporally-structured data. 
Note that for data with no temporal structure LR-MAR will yield a low-dimensional decomposition that will be smoother than for example a PCA decomposition. 
However, no other advantage could in principle be expected from LR-MAR in this case, and we would generally prefer PCA, 
which optimises the amount of explained variance and returns orthogonal components.

\subsection{LR-MAR to aid regression and classification on real data}

Next, we test the efficacy of the approach for classification on electroencephalography (EEG) data
collected across seven subjects, who performed five different activities within five trials.  
The number of sensors (variables) is $N=7$. 
A detailed description of the data can be found in \cite{keirn90}. 
The objective is to discriminate between each pair of tasks.
We use LR-MAR, PCA, ICA and KS with $Q=1$ through 4.
For LR-MAR, the model order $P$ is chosen by using the free energy. 
We take a wavelet time frequency representation \cite{wang04} (using a Morlet wavelet) of the latent components using six frequency scales.
For comparison purposes, we take the same wavelet time frequency representation of the raw channel signals.  
We then run an SVM (endowed with a radial basis function kernel) 
and an adaboost classifier on the Hilbert envelopes of the time frequency representations of both the latent components and the raw data; 
see \cite{hastie08} for some detail about SVM and boosting methodologies. 
Table 1 shows the cross-validated accuracies, where each cross-validation fold corresponds to a different trial. 
LR-MAR outperforms PCA and ICA for both SVM and adaboost classifiers.  $Q=3$ appears to be the best choice. 
The accuracy of the adaboost classifier without a dimensionality reduction step  
is however the highest of all methods. 
SVM, on the other hand, greatly benefits from previous dimensionality reduction.
Even when the results are rather modest in terms of accuracy, 
they still suggest that accounting for time dependencies is useful to produce representative components 
when we deal with strongly temporal data such as EEG (which is known to possess marked oscillatory components),
even when these components might explain less variance from the original data than for example PCA.

\begin{center}
\begin{figure}[h!]
\includegraphics[width=6.0in, totalheight=4.5in]{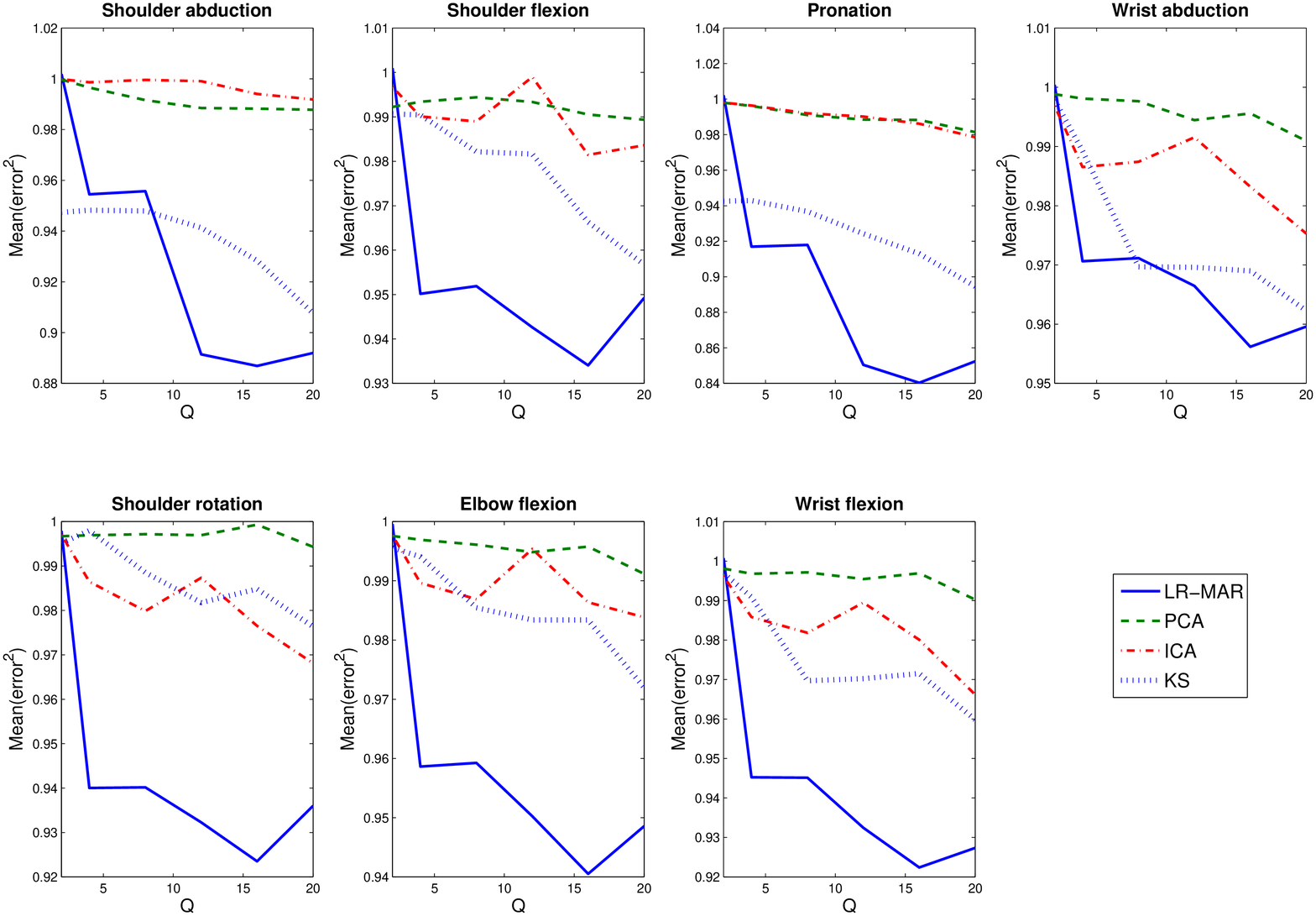}
\vspace{-1.4cm}
\caption{ECoG data: Mean squared error for an ordinary least squares estimation after LR-MAR, PCA, ICA and KS dimensionality reduction, for
different values of $Q$. }
\label{ecog}
\end{figure}
\end{center}

We next deal with a regression problem, 
where the goal is to decode motor outputs from electrocorticogram (ECoG) signals collected in monkeys \cite{chao10}.
There are 11964 time points available, recorded at a sampling rate of 1 kHz. The number of channels is $N=1600$ and the number of outputs to decode is 7. 
We extracted $Q$ latent components using LR-MAR, PCA, ICA and KS with $Q$ ranging between 2 and 20.
For LR-MAR, we limit ourselves to $P=2$ due to the high dimensionality of the data, and assign noninformative values to the hyperparameters. 
Then, we use the extracted latent components as inputs in order to predict the motor responses using ordinary least squares regression.
Figure \ref{ecog} shows the evolution of the mean squared error, computed in a 5-folds cross-validation scheme, as a function of $Q$.
It can be observed that LR-MAR clearly outperforms the rest of the methods, although, for certain outputs (shoulder abduction and pronation), KS can yield lower errors when $Q$ is  low.
In general, all methods' accuracies are similar for $Q=2$, but the improvement of LR-MAR is much more pronounced as $Q$ grows. 
Note that both KS and LR-MAR behave better than PCA and ICA methods for almost all output variables and values of $Q$, suggesting the importance of accounting for time dynamics in this kind of data. 
Remarkably, the best $Q$ for LR-MAR is the same for all responses ($Q=16$). 

Note however that, in a regression context, when the objective is purely predictive, a decomposition that includes information about the response will most likely be more efficient. 
To check this, we have run Bayesian partial least squares (BPLS) \cite{vidaurre13} with $Q$ ranging from 2 to 8 latent components.
BPLS produces an estimation that aims to optimise the prediction power of the model. 
As could be expected, this model thus outperforms the other techniques in terms of accuracy, needing only four components to reach errors that are around $40\%$ lower on average than for example LR-MAR with $Q=8$.
\vspace{1cm}

\begin{table}[b!]
\begin{center}
\caption{EEG data: SVM and adaboost mean classification accuracies over a time-frequency representation of LR-MAR, PCA, ICA and KS components and raw data, averaged over subjects and trials.}
\vspace{0.25cm}
\small
\begin{tabular}{l | lll  lll }
 & \multicolumn{4}{c}{SVM} \\
$Q$  & LR-MAR & PCA &  ICA  & KS    \\
 \hline
$1$ & 0.58 ($\pm$ 0.08  )  &  0.54 ($\pm$ 0.04  )  & 0.53 ($\pm$ 0.10 )  & 0.55 ($\pm$ 0.11)  \\
$2$ & 0.60 ($\pm$  0.10 )  &  0.58 ($\pm$  0.07 )  & 0.56 ($\pm$ 0.10 )  & 0.56 ($\pm$ 0.10 ) \\
$3$ & 0.60 ($\pm$  0.11 )  &  0.57 ($\pm$  0.07 )  & 0.54 ($\pm$ 0.08 )  & 0.56 ($\pm$ 0.09 ) \\  \vspace{-0.25cm}
\end{tabular}
\begin{tabular}{l | lll  lll }
 & \multicolumn{4}{c}{Adaboost} \\
$Q$  & LR-MAR & PCA &  ICA  & KS  \\
 \hline
$1$ & 0.60 ($\pm$ 0.09 )  &  0.57 ($\pm$ 0.06  )  & 0.53 ($\pm$ 0.12  )  & 0.57 ($\pm$ 0.11 ) \\
$2$ & 0.62 ($\pm$ 0.11 )  &  0.60 ($\pm$ 0.10 )  & 0.58 ($\pm$  0.10 )  & 0.57 ($\pm$ 0.11 ) \\
$3$ & 0.64 ($\pm$ 0.12 )  &  0.60 ($\pm$ 0.11  )  & 0.57 ($\pm$ 0.11 )& 0.59 ($\pm$ 0.10 )  \\  \vspace{-0.25cm}
\end{tabular}
\begin{tabular}{ p{0.5\linewidth}   p{0.5\linewidth} }
 \multicolumn{2}{c}{~~~~~~~~~~~~~~~~~~~~~~~~~~~~~~~No dimensionality reduction~~~~~~~~~~~~~~~~~~~~~~~~~~~~~~~~~} \\
~~~~~~~~~~~~~~~~~~~~ SVM ~~~~~~~~~~~~~~~~~~~~~~~~& ~~~~~~~~~~~~~~~~~~~Adaboost ~~~~~~~~~~~~~~~~~~~~~ \\
\hline
~~~~~~~~~~~~~~~~~~~~$0.50 (\pm 0.01)$ ~~~~~~~~~~~~~~~~~~~~~~~ & ~~~~~~~~~~~~~~~~~~ $0.70 (\pm 0.14)$ ~~~~~~~~~~~~~~~~~~~~
\end{tabular}	
\label{best} 
\end{center}
\end{table}
\normalsize

\subsection{wCCA for simultaneous data smoothing and dimensionality reduction}

\begin{center}
\begin{figure}[h!]
\hspace{-0.2cm} 
\includegraphics[width=5.3in, totalheight=3.1in]{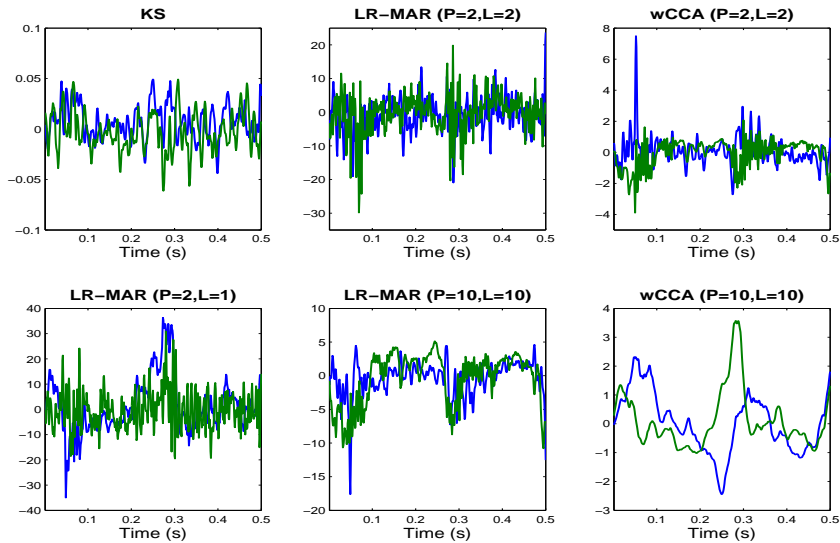}
\vspace{-0.8cm}
\caption{ECoG data: First two latent components for various dimensionality reduction techniques.}
\label{ecog2}
\end{figure}
\end{center}

Finally, we illustrate the smoothing property of wCCA on the same ECoG data using $P=L=2$ and $P=L=10$.
Figure \ref{ecog2} shows the  $Q=2$ extracted components for wCCA compared to KS and LR-MAR with $P=L=2$, $P=L=10$ and $P=2,L=1$.
Note that, whereas the extracted signal has a greater variability for both KS and LR-MAR with $L=1$, 
wCCA exhibits nice smoothing properties for high enough values of $P$ and $L$. 
The extracted signals for LR-MAR with $P=L=2$, $P=L=10$ are less smooth.

\section{Discussion}
\vspace{-0.15cm}

In this paper, we have provided a new factor decomposition for time series that does not assume 
independence of the data points.
Formulated within the Bayesian paradigm, the inference method is able to automatically adjust the complexity of the model and incorporate prior knowledge, if any. 
We have also combined this idea with CCA in order to perform simultaneous smoothing and dimensionality reduction.   

On the one hand, the ARD priors on the $\boldsymbol{W}_i$ matrices can be used to identify which signals are just noise and can thus be excluded from the model, and to prune the order of the autoregressive model.
On the other hand, the ARD prior on the $\boldsymbol{V}$ matrix models the latent data fusion and can indicate the correct latent factor complexity. 
The model can be used as a simplified version of MAR, which needs a large number of parameters when the order and/or the number of signals are high.
Whereas MAR needs $P N^2$ autoregression parameters, LR-MAR entails the estimation of  an effective number of $P N Q $ autoregression coefficients
and hence might scale better in high-dimensional scenarios in terms of statistical efficiency (although not in terms of computation time). 
This way, LR-MAR provides an alternative regularization technique, much as PCA regression is to standard multivariate regression. 

From the MAR representation, one can readily compute connectivity measures from the frequency domain that are of neuroscientific interest, such as coherence or partial coherence \cite{sun04}. 
The same could be achieved from the LR-MAR model using the products $\boldsymbol{W}_i \boldsymbol{V}$.
Because of this direct relation between the frequency representation and the MAR/LR-MAR coefficients, 
our approach is expected to obtain more meaningful components than other techniques  if we aim to model oscillatory data.
This is for example the case of neuronal population dynamics in the brain;
see for example \cite{jacobs07} and references therein.
Note that the elicitation of the frequency characteristics is not that straightforward for an KS model.
Besides its simplicity and computational advantages, this is admittedly an advantage of  LR-MAR over KS. 

Of course, one can always use standard PCA or ICA over a time frequency representation of the data or take a time frequency representation of the PCA/ICA components.
In this case, the settings that govern the time frequency decomposition (e.g. using wavelets) is separetely done from the PCA/ICA dimensionality reduction step and is 
more or less heuristically determined according to some subjective belief or empirical evaluation.
The same claim can be done for the use of functional data analysis \cite{ramsay05} over PCA components, where temporal smoothness is enforced on the components. 
Contrary to these two-step procedures, LR-MAR is a more compact approach that can make use of the Bayesian machinery for tuning the parameters of the model. 

The most relevant parameters to select are $Q$ and $P$, 
although, as mentioned, the use of ARD priors mitigates the impact of this choice. 
Hence, starting with relatively high values, the estimates values of $\boldsymbol\alpha^{-2}$ and $\boldsymbol\gamma^{-2}$
can provide an idea of the proper values to use. 
However, if computational time is not problematic, model selection based on the free energy is recommended.
Nonparametric inference of these parameters is definitely a possible route as well, 
but the model loses the conjugacy and variational inference is no longer possible. 
Although not performed in this paper for simplicity, the variational inference of the prior hyperparameters can be also carried out. 

We have tested the model on both synthetic and real neuroscience time series data, 
proving empirically a good performance in explaining the variance with a minimum number of components
and in providing a meaningful representation for a subsequent supervised learning step. 
In this case, these experiments depart from a traditional regression/classification paradigm, where the testing data cannot be at all used in the training procedure.
Instead, we are here using the entire data set to perform the dimensionality reduction step, so that the testing data intervene in the dimensionality reduction step for the training data and vice versa.
In many applications, however, we would typically be given an entire signal and a response (either continuous or categorical) for only part of the signal, and would then be asked to give a prediction for the missing responses. 
It is thus fair to use the entire signal to find a good representation for the subsequent prediction as far as we do not use information about the response of the testing data. 
We could rephrase this as a kind of data completion problem where the missing data are the actual responses. 
Note that  this would not be the case for applications where a quick online prediction is required, such a brain-computer interface application.
In this case, it would probably be too computationally inefficient to repeat the dimensionality reduction step each time a new batch of data is presented. 
The logical alternative would be to use the parameters of the dimensionality reduction step obtained from training data on each new batch of data without continuously refitting the model. 

A further extension of the model is to provide a mechanism to account for changes in the time series dynamics, 
allowing $\boldsymbol{W}$ and $\boldsymbol{V}$ to depend on a hidden state variable.
This way, for each state, we would have different distributions for matrices $\boldsymbol{W}$ and $\boldsymbol{V}$ 
(and for $\boldsymbol\alpha^{-2}$ and $\boldsymbol\gamma^{-2}$). 
The hidden state variables would be modelled using the Markov assumption, so that we could use the variational Baum-Welch recursions (see e.g. \cite{ghahramani00,redek05}) to make inference on them.
The model could be then used for unsupervised classification.
The above equations can be adapted for this purpose without much difficulty. 

%



\section{Appendix A: Computation of the free energy}

The variational free energy is given by 
\begin{eqnarray}
\label{fe}
\mathcal{F} =   
\int F(\boldsymbol{Z}) \log F(\boldsymbol{Z}) \, d \boldsymbol{Z} + \int F(\boldsymbol\Phi) \log \frac{F(\boldsymbol\Phi)}{Pr(\boldsymbol\Phi)} \, d \boldsymbol\Phi  \quad\quad\quad\quad\quad\quad\quad\quad\quad\\
\nonumber - \int  F(\boldsymbol{Z}) F(\boldsymbol\Phi) \log P(\boldsymbol{Y} | \boldsymbol{Z}, \boldsymbol\Phi) \, d \boldsymbol{Z} \, d \boldsymbol\Phi  
- \int  F(\boldsymbol{Z}) F(\boldsymbol\Phi) \log P(\boldsymbol{Z} | \boldsymbol\Phi) \,   d \boldsymbol{Z} \, d \boldsymbol\Phi,
\end{eqnarray}

\noindent 
where $\boldsymbol\Phi$ denotes all the model parameters. 
The first term is the negative entropy of $\boldsymbol{Z}$, which 
can be readily computed as 

\begin{equation*}
\int F(\boldsymbol{Z}) \log F(\boldsymbol{Z}) \, d \boldsymbol{Z} = 0.5 \sum_{t=P+1}^T \log | \,2 \pi e \boldsymbol{S}_{\boldsymbol{z}_t} |.
\end{equation*}

\noindent
The second term represents the Kullback-Leibler divergences between the prior and approximate posterior distributions.
The last two terms define the average log-likelihood of the model,  

\begin{eqnarray*}
- \int  F(\boldsymbol{Z}) F(\boldsymbol\Phi) \log P(\boldsymbol{Z} | \boldsymbol\Phi) \,   d \boldsymbol{Z} \, d \boldsymbol\Phi = 
\quad\quad\quad\quad\quad\quad\quad\quad\quad\quad\quad\quad\quad\quad\quad\,\,\\
\frac{T-P}{2} \log(2\pi)^{Q} + \frac{1}{2} \, \mathrm{tr} \Big( \boldsymbol{E}'_{\boldsymbol{Z}} \boldsymbol{E}_{\boldsymbol{Z}} \Big) + \frac{1}{2} \sum_{t=P+1}^T\mathrm{tr} \Big( S_{\boldsymbol{z}_t} \Big) 
+ \frac{1}{2} \sum_{j=1}^Q  \mathrm{tr} \Big(\boldsymbol{Y}^{-} \boldsymbol{S}_{\boldsymbol{w}_j} \boldsymbol{Y}^{-'} \Big)  , 
\end{eqnarray*}
\noindent
and
\begin{eqnarray*}
- \int  F(\boldsymbol{Z}) F(\boldsymbol\Phi) \log P(\boldsymbol{Y} \big{|} \boldsymbol{Z}, \boldsymbol\Phi) \, d \boldsymbol{Z} \, d \boldsymbol\Phi = 
\quad\quad\quad\quad\quad\quad\quad\quad\quad\quad\quad\quad\quad\quad \,\,\\
\frac{T-P}{2} \log(2\pi)^{N} + \frac{T-P}{2} \sum_{n=1}^N \Psi\Big(\frac{\bar{\iota}+1-n}{2}\Big) 
- \frac{T-P}{2} \sum_{n=1}^N \log \bar{a}_n \\
+ \frac{1}{2} \mathrm{tr} \Big( \boldsymbol{E}'_{\boldsymbol{Y}} \bar{\boldsymbol\Omega}^{-1} \boldsymbol{E}_{\boldsymbol{Y}}  \Big)
+ \frac{1}{2} \sum_{n=1}^N \mathrm{tr} \Big(  \bar{\boldsymbol{Z}} \boldsymbol{S}_{\boldsymbol{v}_n}   \bar{\boldsymbol{Z}}'   \Big) 
+ \frac{1}{2}\sum_{t=P+1}^T  \sum_{n=1}^N   \mathrm{tr} \Big(  \bar{\boldsymbol{v}}_n \bar{\boldsymbol{v}}'_n S_{\boldsymbol{z}_t} +   \boldsymbol{S}_{\boldsymbol{v}_n}    S_{\boldsymbol{z}_t}  \Big) 
\end{eqnarray*}

\noindent
where $\Psi(\cdot)$ is the digamma function, 
$\boldsymbol{E}_{\boldsymbol{Z}} = \bar{\boldsymbol{Z}} - \boldsymbol{Y}^{-}  \bar{\boldsymbol{W}}$ and
$\boldsymbol{E}_{\boldsymbol{Y}} = \boldsymbol{Y}^{+} - \bar{\boldsymbol{Z}}  \bar{\boldsymbol{V}}$.

\vspace{1cm}
\section*{Bibliography}

\bibliographystyle{unsrt}
\bibliography{lowrank}

\end{document}